\documentclass[german,english]{bvm2018}

\title{Motion Artifact Detection in Confocal Laser Endomicroscopy Images}

\author{Maike~Stoeve$^1$, Marc~Aubreville$^1$, Nicolai~Oetter$^{2,3}$, Christian~Knipfer$^{4,3}$, Helmut~Neumann$^{5,3}$, Florian~Stelzle$^{2,3}$, Andreas~Maier$^{1,3}$}

\authorrunning{Stoeve et al.}
\institute{%
$^1$Pattern Recognition Lab, Computer Science, Friedrich-Alexander-Universit{\"a}t Erlangen-N{\"u}rnberg\newline
$^2$Department of Oral and Maxillofacial Surgery, University Hospital Erlangen, Friedrich-Alexander-Universit{\"a}t Erlangen-N{\"u}rnberg\newline
$^3$Erlangen Graduate School in Advanced Optical Technologies (SAOT), Friedrich-Alexander-Universit{\"a}t Erlangen-N{\"u}rnberg\newline
$^4$Department of Oral and Maxillofacial Surgery, University Medical Center Hamburg-Eppendorf\newline
$^5$First Department of Internal Medicine, University Hospital Mainz, Johannes Gutenberg-Universit{\"a}t Mainz}

\email{maike.stoeve@fau.de}

\begin{document}

%==============================================================================
% w?hlen Sie mit dem Befehl \selectlanguage die Sprache aus, in der Ihr 
% Proceeding verfasst ist
%
%\selectlanguage{german}
\selectlanguage{english}

\maketitle

\begin{abstract}
Confocal Laser Endomicroscopy (CLE), an optical imaging technique allowing non-invasive examination of the mucosa on a (sub)- cellular level, has proven to be a valuable diagnostic tool in gastroenterology and shows promising results in various anatomical regions including the oral cavity. Recently, the feasibility of automatic carcinoma detection for CLE images of sufficient quality was shown. However, in real world data sets a high amount of CLE images is corrupted by artifacts. Amongst the most prevalent artifact types are motion-induced image deteriorations. In the scope of this work, algorithmic approaches for the automatic detection of motion artifact-tainted image regions were developed. Hence, this work provides an important step towards clinical applicability of automatic carcinoma detection. Both, conventional machine learning and novel, deep learning-based approaches were assessed. The deep learning-based approach outperforms the conventional approaches, attaining an AUC of 0.90.
\end{abstract}

\section{Introduction}
With over 500,000 diagnosed cases each year, head and neck squamous cell carcinoma (HNSCC) is considered the sixth most common cancer type worldwide \cite{2567-01}. For diagnosis of HNSCC, invasive biopsy and subsequent histopathological examination is applied as gold standard method \cite{2567-02}. As alternative, non-invasive optical imaging technologies as narrow-band imaging, autofluorescence imaging and confocal laser endomicroscopy (CLE) are gaining interest in research \cite{2567-03}. Among those technologies, CLE has already proven to be a valuable diagnostic tool in gastroenterology and has been successfully applied to examine lesions in the oral cavity \cite{2567-04}. Both, the interpretation of CLE images and the histopathological examination of tissue samples require experience and proficiency and entail a subjective component \cite{2567-05}. Thus, promising approaches using machine learning techniques for automatic carcinoma detection based on CLE images were developed \cite{2567-06}.

As the accuracy of these algorithms is highly affected by the occurrence of artifacts, artifact-tainted images were excluded manually prior to training in a time-consuming manual labeling step in all known approaches. In the scope of this work, an automatic motion artifact detection pipeline was developed and evaluated. Hence, this work provides the basis to integrate a fully automatic motion artifact detection into existing carcinoma detection frameworks, an important step towards clinical applicability.

\subsection{Motion artifacts}
\begin{figure}[htb]
	\setlength{\figbreite}{0.3\textwidth}
	\centering
	\subfigure[]{\includegraphics[width=30mm]{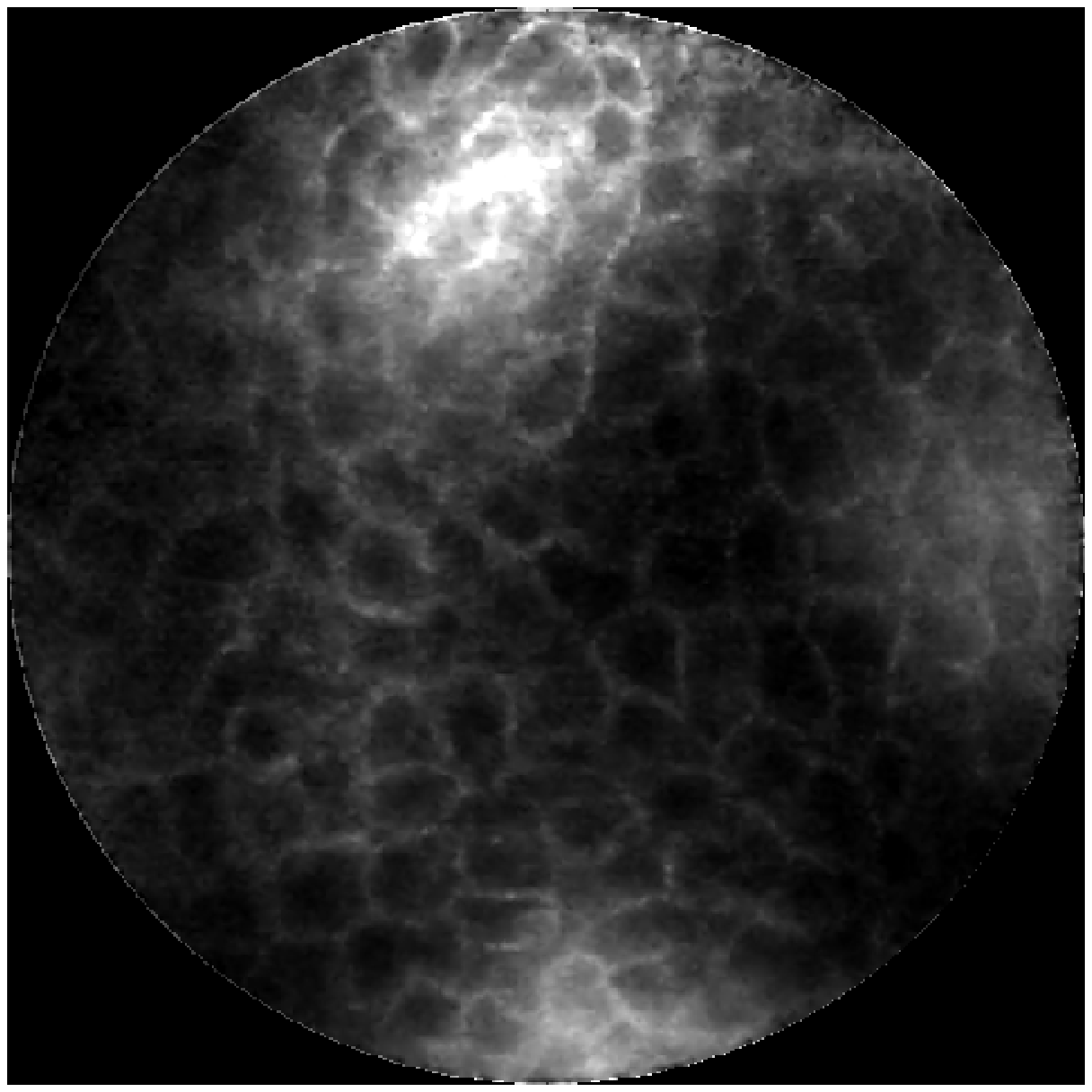}}
	\subfigure[]{\includegraphics[width=30mm]{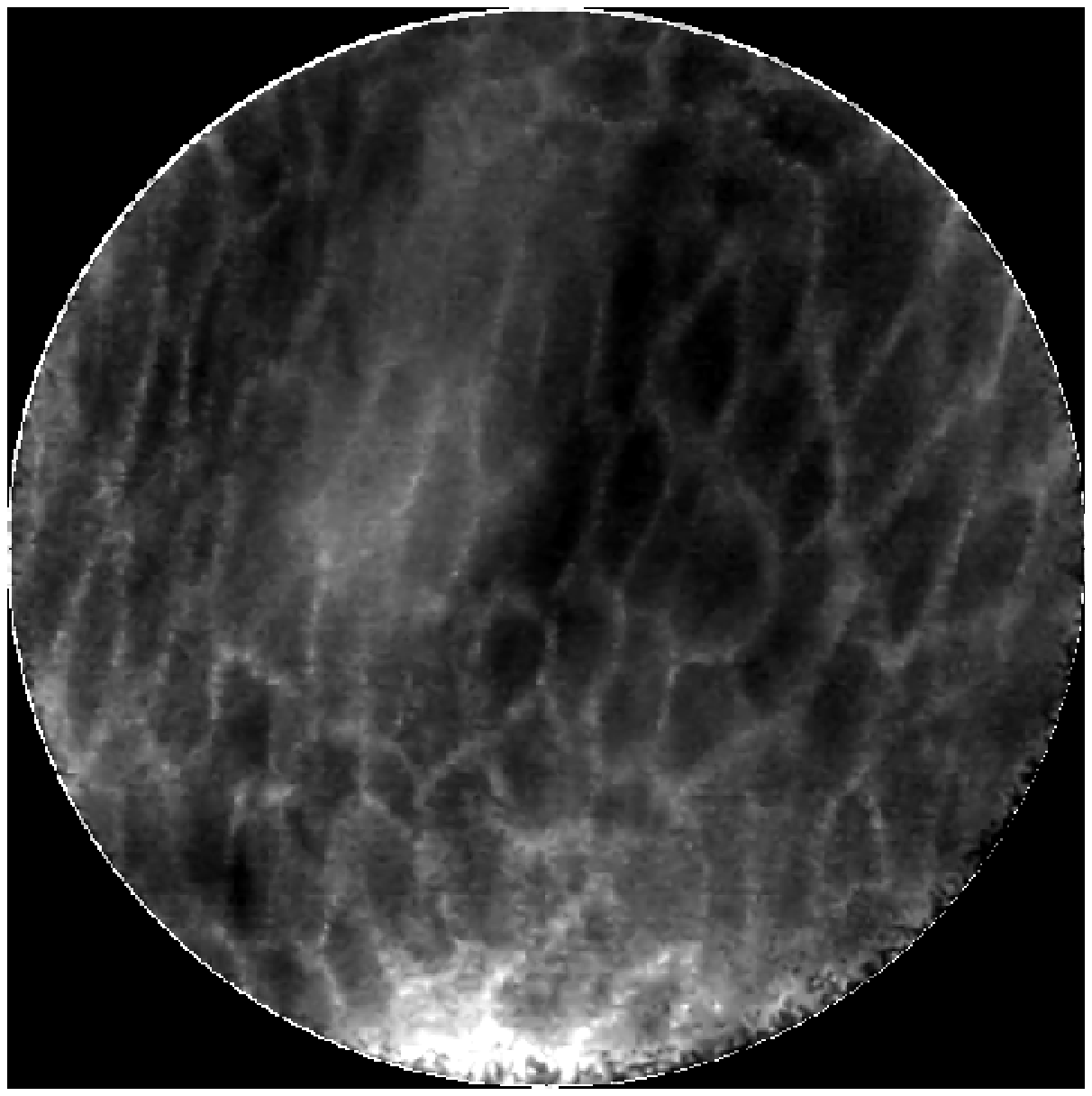}}
	\subfigure[]{\includegraphics[width=30mm]{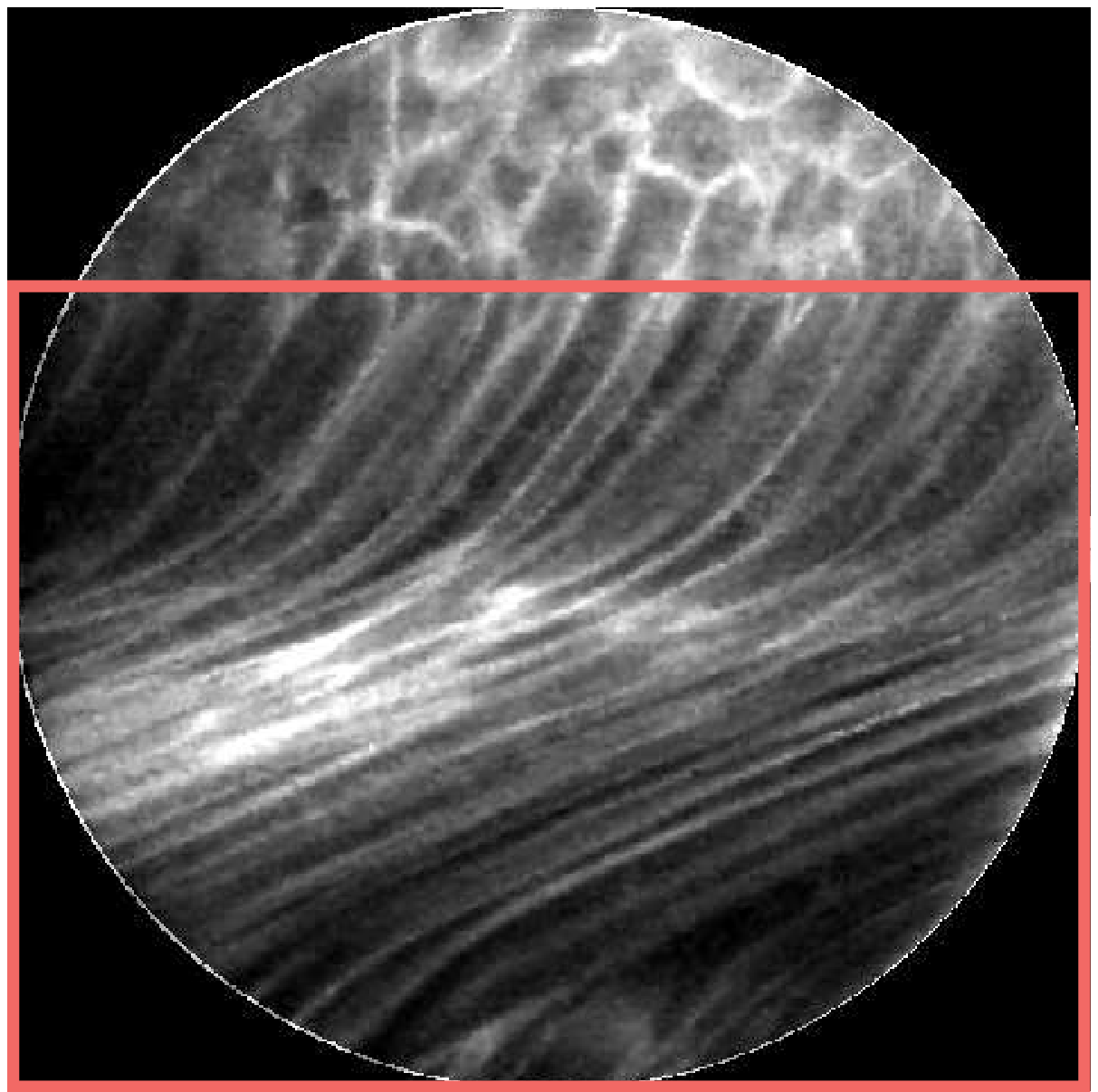}}
	\caption{CLE images containing (a) no artifacts or motion artifacts manifesting in (b) stretched cells or (c) stripe patterns (region marked in red).}
	\label{fig:2567-fig1-3}
\end{figure}

A frequent cause of image impairment are motion artifacts. They can either be caused by movements of the investigated anatomical structures or motion induced by the physician. The proportion of motion artifact images compared to good quality images is highly dependent on the experience of the physician \cite{2567-07}. As shown in Fig.~\ref{fig:2567-fig1-3}, two different manifestations can be observed. The first manifestation is characterized by streaky patterns originating from the repeated acquisition of the same, shifted image line. This pattern can be observed, when the sum of the velocities of probe, organ and scanning is approximately equal to zero. This requires an organ movement in the same direction as the sampling pattern or a probe motion in the correspondingly opposite direction. If a significant relative motion results from organ or probe motion, the cells are stretched or blurred. This second manifestation is hard to detect, as distinction between elongated cells is difficult without the observation of adjacent frames. Finally, motion artifacts can impair the whole image leading to a total loss of information or only influence parts of the image still allowing a diagnosis based on untainted regions \cite{2567-07}. Due to the meander-shaped optical sampling pattern, only whole image rows are affected by motion artifacts.

\section{Materials and methods}
For the present work, 116 CLE sequences comprising 11,234 images from 12 patients and 4 sites within the oral cavity, namely the upper alveolar ridge, the lower inner labium, the palatal region and the lesion site itself were utilized. The CLE sequences were recorded at the department of Oral and Maxillofacial Surgery of the University Hospital Erlangen by a standalone probe-based CLE system (Cellvizio, Mauna Kea Technologies, Paris, France). The obtained images are approximately of size 576$\times$578 pixels. The overall image quality was assessed by an expert and artifact regions were annotated manually within each image. Details on the data can be found in \cite{2567-06}. 

In the scope of this work, two different methodical approaches were established for the detection of motion artifact-tainted CLE image regions. The first approach uses conventional pattern recognition methods extracting characteristic image features, whereas the second approach applies deep learning strategies. For both approaches, images with low signal-to-noise ratio are excluded. The remaining CLE images are converted to 8-bit integer values after a quantile-based dynamic range compression following Aubreville et al. \cite{2567-06} was performed. 

\subsection{Feature-based motion artifact detection}
The feature-based approach consists of three steps: pre-processing, feature extraction and classification. The pre-processing step is required due to the unusual round shape of CLE images complicating the feature extraction process. Jaremenko et al. circumvented this problem by dividing the image in overlapping, square patches and concatenating the information of all patches for the classification of the whole image \cite{2567-08}. As motion artifacts always cover the whole width of an image, slices with a width of the maximum extent of the CLE image in x-direction, a fixed height of 128 pixels and an overlap of 30\ts \% (HOG) or 50\ts \% (corrAngle) are extracted. To allow the detection of motion artifact-tainted slices of an image and differentiate them from untainted slices of the same image, either Histogram of Oriented Gradient (HOG) or angle of maximum correlation (corrAngle) features are extracted for each slice. For classification, a random forest classifier (RF) and a support vector machine (SVM) classifier were used. Undersampling of the majority class is applied to deal with the class imbalance. Prior to training of the linear SVM classifier, the feature vector is standardized. For evaluation purposes of both classifiers, 5-fold cross-validation is used. The different slices of one image are all assigned to the same fold.

\subsubsection{Histogram of oriented gradients (HOG)}
The Histogram of Oriented Gradient (HOG) feature descriptor of Dalal and Triggs is frequently used in the field of object recognition \cite{2567-09}. Basically, the occurrence of gradient directions is computed in a local image region characterizing the local shape of an object. In contrast to the original pipeline, no gamma and color normalization is used. A cell size of 32$\times$32 and block size of 64$\times$64 resulted in the best classification performance. Due to the varying length of CLE image slices, the obtained HOG feature vectors are of varying length. Hence, mean, standard deviation, skewness and kurtosis are computed over all 9 bins of the feature vector individually. The statistical properties for all bins are subsequently concatenated to form a feature vector of a fixed length of 36.

\subsubsection{Angle of Maximum Correlation (corrAngle)}
The angle of maximum correlation feature (corrAngle) is designed under consideration of the origin of the streaky patterns visible in most motion artifact-tainted images. The direction of the relative motion accounts for the angle $\theta$ characterizing the direction of the stripe pattern. To create a feature vector describing the presence or absence of motion artifact patterns in an image slice with length $L$, a centered reference segment of row $i$ of the CLE image slice is extracted. For a set of angles equally distributed between $\frac{\pi}{8}$ and $\frac{7}{8} \pi$ and a fixed radius $R$, comparative segments are extracted with center at $(\frac{L}{2} + R \sin(\theta), j + R + R \cos(\theta))$ using bilinear interpolation. Then, the correlation coefficient of the reference segment and each comparative segment is computed as measurement of similarity. The angle responsible for the highest correlation coefficient is saved. This procedure is repeated for each row despite the first and last $R$ rows. Finally, all approximated motion angles are concatenated to form the final feature vector. For a motion artifact-tainted image slice, the approximated angle is constant (stripe pattern). In contrast, CLE image slices without motion artifacts show high variations of the approximated motion angle over image rows.

\subsection{Deep learning-based motion artifact detection (artiNet)}
\begin{figure}[t]
\centering
\caption{Visualization of the artiNet for motion artifact detection in CLE images based on an Inception v3 network \cite{2567-10} pre-trained on ImageNet.}
\includegraphics[width=115mm]{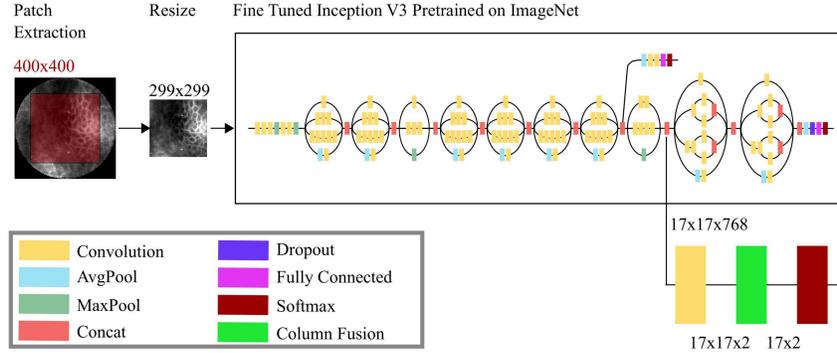}
\label{fig:2567-fig4}
\end{figure}

Building on the Inception v3 architecture of Szegedy \cite{2567-10} pretrained on \mbox{ImageNet}, the deep learning-based detector of motion artifact-tainted CLE image areas (artiNet) is build by inserting a 2d 1$\times$1 convolutional layer after the eighth inception block followed by a column fusion and a softmax layer as depicted in Fig.~\ref{fig:2567-fig4}. Thus, the input representation for the new layers still entails spatial information. The convolutional layer is used to map the 17$\times$17$\times$768 dimensional input to the two output classes motion artifact and good quality. The result is a 17$\times$17 grid of predictions. To obtain predictions of motion artifact regions covering the whole image width comparable to the slices of the conventional approach, a column fusion layer extracts the maximum over the image width resulting in a 17$\times$2 representation fed to the final softmax layer.

Prior to training the grayscale images are transformed to RGB color representation. Then, a centered image patch of 400$\times$400 pixels is extracted and resized to 299$\times$299 pixels. Due to the slice-wise detection of motion artifacts and available corresponding labels effectively representing a fully convolutional approach with a network capacity according to one with smaller patches, data augmentation seems less important. Within the TensorFlow framework, the \mbox{Inception} network was fine-tuned in $2000$ training steps using the Adam optimizer with an initial learning rate of $5\cdot10^{-6}$. For the training of the new layers, a learning rate of $5\cdot10^{-5}$ was used. In each training step, a minibatch consisting of 25 randomly selected instances of both classes is processed to deal with the class imbalance of the data set. For evaluation, a leave-one-patient-out cross-validation was performed.

\section{Results}
\begin{figure}[t]
\caption{Results of motion artifact detection in CLE images.}
\centering
\subfigure[ROC curves visualizing the average cross-validation performance of the motion artifact detection.]{\includegraphics[width=60mm]{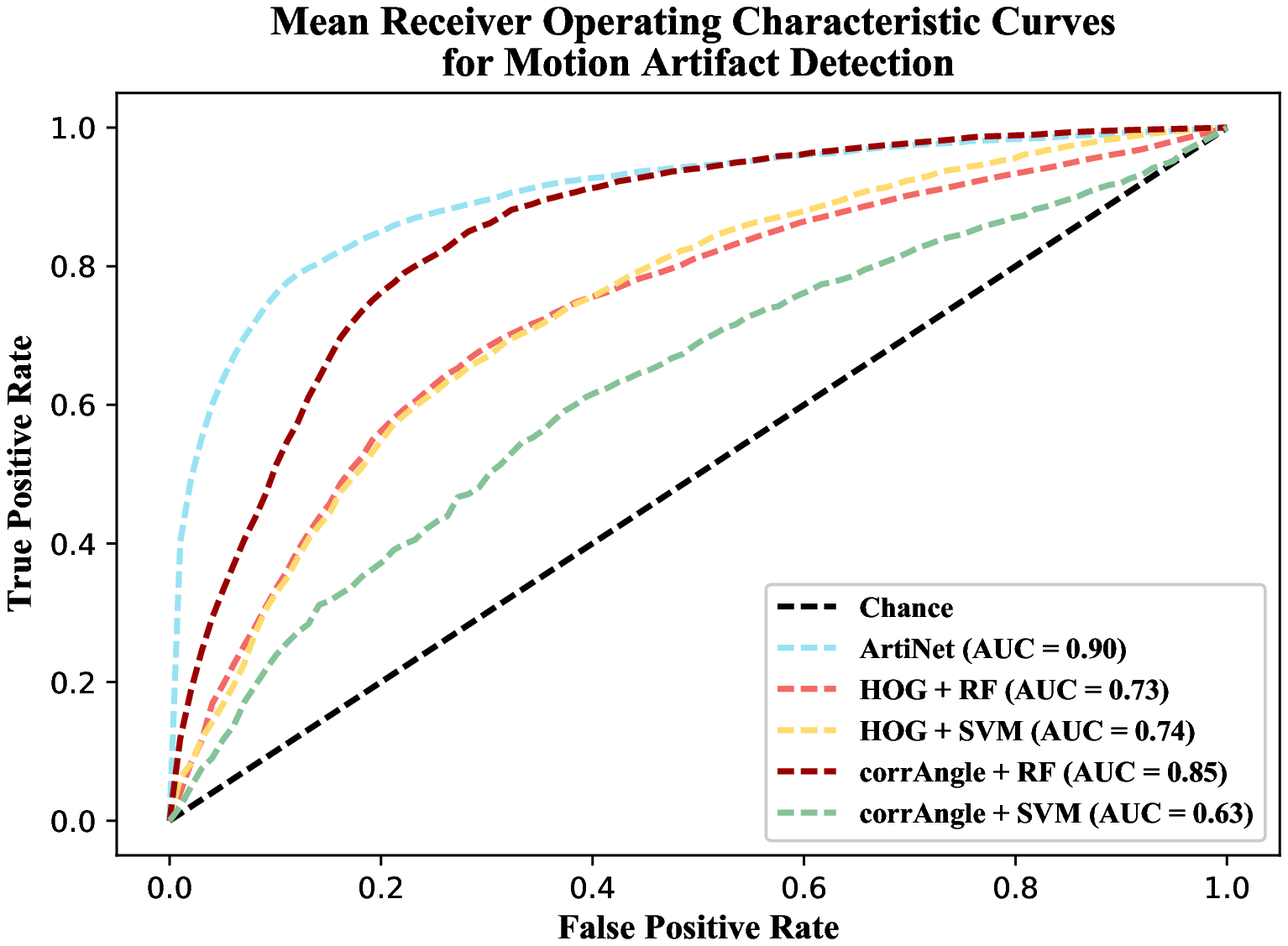}}
\hfill
\subfigure[Example images showing slicewise predictions of the artiNet compared to class labels.]{\includegraphics[width=40mm]{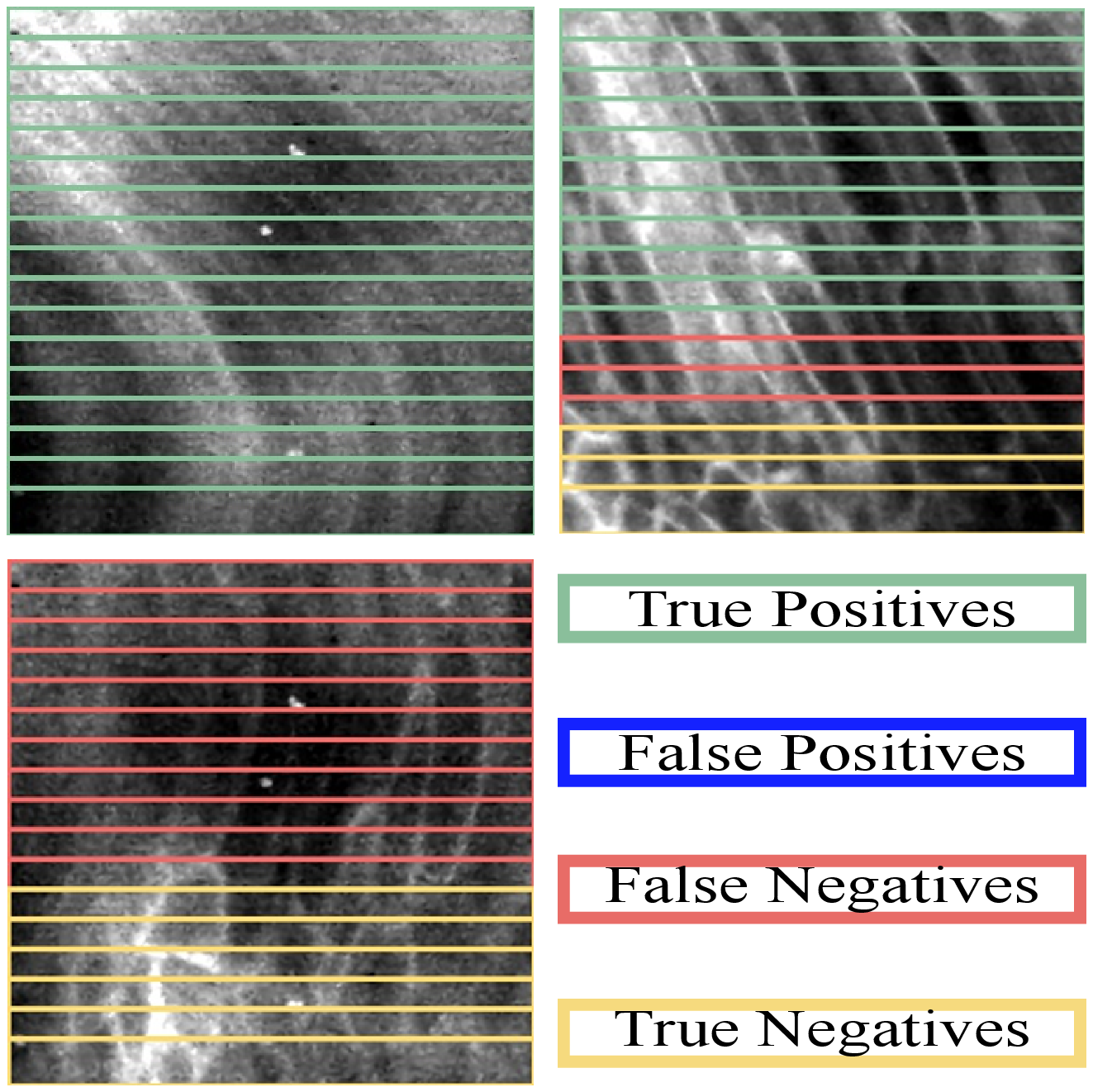}}
\label{fig:2567-fig5-6}
\end{figure}

The top result is obtained by the artiNet, where an AUC of 0.90 is achieved. The best performance following the conventional pattern recognition pipeline is achieved by the corrAngle feature and RF classification. For this combination, an AUC of 0.85 is reached. The corrAngle-based motion detection with RF performs significantly better than the RF and SVM predictors trained on HOG, where approximately equal ROC curves with AUC values of 0.73 and 0.74 are achieved. In contrast to the RF classification performance, the results of the SVM classifier trained on the corrAngle features are poor. For additional insight into the performance of the artiNet, the comparison of predictions for single image slices and respective labels is visualized in Fig.~\ref{fig:2567-fig5-6}.

\section{Discussion}
Due to the included patch extraction step of the artiNet, the image borders are discarded. For motion artifact detection, the center of the slice is representative as a motion induced deterioration covers the whole extent in x-direction. Still, the information at the top and bottom of the image is removed. Thus, motion artifacts only deteriorating the rejected areas can not be detected. An improved performance is to be expected if the whole image is used. As the manifestation of stretched cells is underrepresented in the data set, a performance gap between the two possible manifestations of motion artifacts, stripe patterns and stretched cells might occur. Moreover, the performance of the proposed methods was only assessed on data of one clinical team. Hence, additional training instances are required to obtain a robust artifact detector. 
\bibliographystyle{bvm2018}

\end{document}